\documentclass[journal]{IEEEtran}
\usepackage{graphicx}
\usepackage{epsfig}

\usepackage{amsmath,amssymb,amsfonts}
\usepackage{amsbsy}
\usepackage{graphicx}
\usepackage{graphics}
\usepackage{algorithm}
\usepackage[noend]{algorithmic}
\usepackage{subfigure}
\usepackage{cite}
\usepackage{slashbox}
\usepackage{multirow}
\usepackage{pdfsync}
\usepackage{epstopdf}
\usepackage{color}
\usepackage{amsthm}
\usepackage{boxedminipage}
\usepackage{textcomp}

\newcommand{\Rbb}{\mathbb{R}}

\newcommand{\G}{{\mathcal{G}}}
\newcommand{\E}{{\mathcal{E}}}

\newcommand{\V}{{\mathcal{V}}}

\renewcommand{\L}{{\mathcal{L}}}
 
 \DeclareMathOperator*{\argmin}{argmin}

\ifCLASSINFOpdf
 
\else
  
\fi

\begin{document}

\title{Graph-based compression of  dynamic \\ 3D point cloud sequences}
\author{Dorina~Thanou, 
        Philip A. Chou, 
        and Pascal~Frossard       
 \thanks{D. Thanou and P. Frossard are with Ecole Polytechnique F\'ed\'erale de Lausanne (EPFL), Signal Processing Laboratory-LTS4,  Lausanne, Switzerland. P. A. Chou is with Microsoft Research, Redmond, WA, USA.  ({e-mail:\{dorina.thanou, pascal.frossard\}@epfl.ch}, pachou@microsoft.com).}
 \thanks{Part of this work has been accepted for presentation in the International Conference on Image Processing (ICIP), Quebec, Canada, September 2015.}}
 
\maketitle
\maketitle

\begin{abstract}
This paper addresses the problem of compression of 3D point cloud sequences that are characterized by moving 3D positions and color attributes. As temporally successive point cloud frames are similar, motion estimation is key to effective compression of these sequences. It however remains a challenging problem as the point cloud frames have varying numbers of points without explicit correspondence information. We represent the time-varying geometry of these sequences with a set of graphs, and consider 3D positions and color attributes of the points clouds as signals on the vertices of the graphs. We then cast motion estimation as a feature matching problem between successive graphs. The motion is estimated on a sparse set of representative vertices using new spectral graph wavelet descriptors. A dense motion field is eventually interpolated by solving a graph-based regularization problem. The estimated motion is finally used for removing the temporal redundancy  in the predictive coding of the 3D positions and the color characteristics of the point cloud sequences. Experimental results demonstrate that our method is able to accurately estimate the motion between consecutive frames. Moreover, motion estimation is shown to bring significant improvement in terms of the overall  compression performance of the sequence. To the best of our knowledge, this is the first paper that exploits both the spatial correlation inside each frame (through the graph) and the temporal correlation between the frames (through the motion estimation) to compress the color and the geometry of 3D point cloud sequences in an efficient way. 
\end{abstract}

\IEEEpeerreviewmaketitle
\section{Introduction}
\label{sec:intro}

\begin{figure}[t]
      \centering
         
          { \includegraphics[width=7cm]{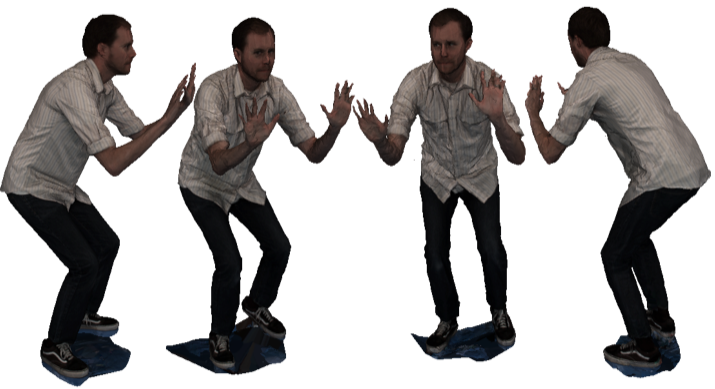} } 
  
           \caption{Example of a sequence of frames, captured at different time instances } 
        \label{fig:sequence_example}
\vspace{-0.1cm}
\end{figure}

Dynamic 3D scenes such as humans in motion can now be captured by arrays of color plus depth (or `RGBD') video cameras \cite{Loop_2013}, and such data is getting very popular in emerging applications such as animation, gaming, virtual reality, and immersive communications.  The geometry captured by RGBD camera arrays, unlike computer-generated geometry, has little explicit spatio-temporal structure, and is often represented by sequences of colored point clouds. Frames, which are the point clouds captured at a given time instant as shown in Fig.~\ref{fig:sequence_example},   may have different numbers of points, and there is no explicit  association between points over time.  Performing motion estimation, motion compensation, and effective compression of such data is therefore a challenging task.

In this paper, we focus on the compression of the 3D geometry  and color attributes and propose a novel motion estimation and compensation scheme that exploits temporal correlation in sequences of point clouds.  To deal with the large size of these sequences, we consider that the point clouds are voxalized,  that is, their 3D positions are quantized to a regular,
axis-aligned, 3D grid having a given stepsize.  This quantization of the space is commonly achieved by modeling the 3D point cloud sequences as a series of octree data structures \cite{Loop_2013}, \cite{Huang2008}, \cite{schnabel_2006}. In contrast to polygonal mesh representations, the octree structure exploits the spatial organization of the 3D points, which results in easy manipulations and allows real-time processing.  In more details,  an octree is a tree structure with a predefined depth, where every branch node represents a certain cube volume in the 3D space, which is called a voxel.  
A voxel containing a point is said to be occupied.  Although the overall voxel set lies in a regular grid, the set of occupied voxels are non-uniformly distributed in space.   To uncover the irregular structure of the occupied voxels inside each frame,  we consider
points as vertices in a graph $\mathcal{G}$, with edges between nearby vertices.  
Attributes of each point $n$, including 3D position $p(n)=[x,y,z](n)$ and color components $c(n)=[r,g,b](n)$, are treated as signals residing on the vertices of the graph.
As frames in the 3D point cloud sequences are correlated, the graph signals at consecutive time instants are also correlated. Hence, removing temporal correlation  implies  comparing the behavior of the signals residing on the vertices of consecutive graphs. The estimation of the correlation is however a challenging task as the graphs usually have different number of nodes and no explicit correspondence information between the nodes is available in the sequence.

We build on our previous work \cite{Thanou2015}, and  propose a novel algorithm for motion estimation and compensation in 3D point cloud sequences. We cast motion estimation as a feature matching problem on dynamic graphs. In particular, we compute new local features at different scales with spectral graph wavelets (SGW) \cite{Hammond2010} for each node of the graph. 
Our feature descriptors, which consist of the wavelet coefficients of each of the signals placed in the corresponding vertex, are 
then used to compute point-to-point correspondences between graphs of different frames. We  match our SGW features in different graphs with a criterion that is based on the Mahalanobis distance and trained from the data. To avoid inaccurate matches, we first compute the motion on a sparse set of matching nodes that satisfy the matching criterion. We then  interpolate the motion of the other nodes of the graph by solving a new graph-based quadratic regularization problem, which promotes smoothness of the motion vectors on the graph in order to build a consistent motion field.

\begin{figure}[t]
      \centering
         
          { \includegraphics[width=9cm]{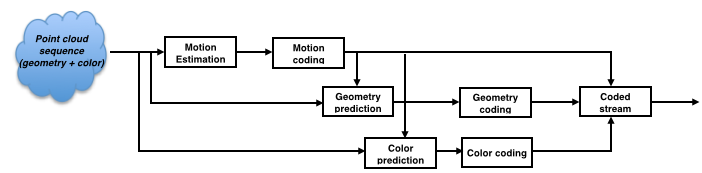} \label{figure_intro}} 
  
           \caption{Schematic overview of the encoding architecture of a point cloud sequence. Motion estimation is used to reduce the temporal redundancy for efficient  compression of  the 3D geometry and  the color attributes. } 
        \label{fig:figure_intro}
\vspace{-0.1cm}
\end{figure}

Then, we design a compression system for 3D point cloud sequences, where we exploit the estimated motion information in the predictive coding of the geometry and color information.  The basic blocks of our compression architecture are shown in Fig.~\ref{fig:figure_intro}. We  code the motion field in the graph Fourier domain by exploiting its smoothness on the graph.  Temporal redundancy in consecutive 3D positions is removed by coding the structural difference between the  target frame and the motion compensated reference frame. The structural difference is efficiently described in a binary stream format  as described in \cite{Kammerl2012}. Finally, we  predict the color of the target frame by interpolating it from the color of the motion compensated reference frame.    Only the difference between the actual color information and the result of the motion compensation is actually coded with a  state-of-the-art encoder for static octree data  \cite{Zhang2014}.  Experimental results  illustrate that our motion estimation scheme is efficient as it can capture the motion between consecutive frames.  Moreover,  introducing motion compensation in compression of 3D point cloud sequences  results in significant improvement in terms of rate-distortion  performance of the overall system, and in particular in the compression of the color attributes where we achieve a gain of up to 10 dB in comparison to state-of-the-art encoders. The contribution of the paper is summarized as follows.  To the best of our knowledge, the proposed encoder is the first one in the existing literature that exploits motion estimation to remove the temporal redundancy for efficient coding of point cloud sequences, without going first through the expensive conversion process into a temporally consistent polygonal mesh. Second, we represent the point cloud sequences as a set of graphs and we solve the motion estimation problem as a feature matching problem in dynamic graphs. Third,  we propose a differential coding scheme for color compression that provides significant gain in terms of coding performance. 

The rest of the paper is organized as follows. First, in Section  \ref{sec:related_work}, we review the existing work in the literature that studies the problem of  compression of 3D point clouds.  Next, in Section \ref{sec:representation_with_graphs}, we  describe the representation of 3D point clouds by performing an octree decomposition of the 3D space and we introduce graphs to capture the irregular structure of this representation. 
 The motion estimation  scheme is presented  in Section \ref{sec:Motion_estimation_compensation}. The estimated motion is then applied to the predictive coding of the geometry and the color in Section \ref{sec:coding_of_3D_sequences}.  Finally, experimental results are given in Section \ref{sec:experimental_results}.

\begin{figure*}[th]
      \centering
            \subfigure[Original point cloud]{ \includegraphics[width=4cm]{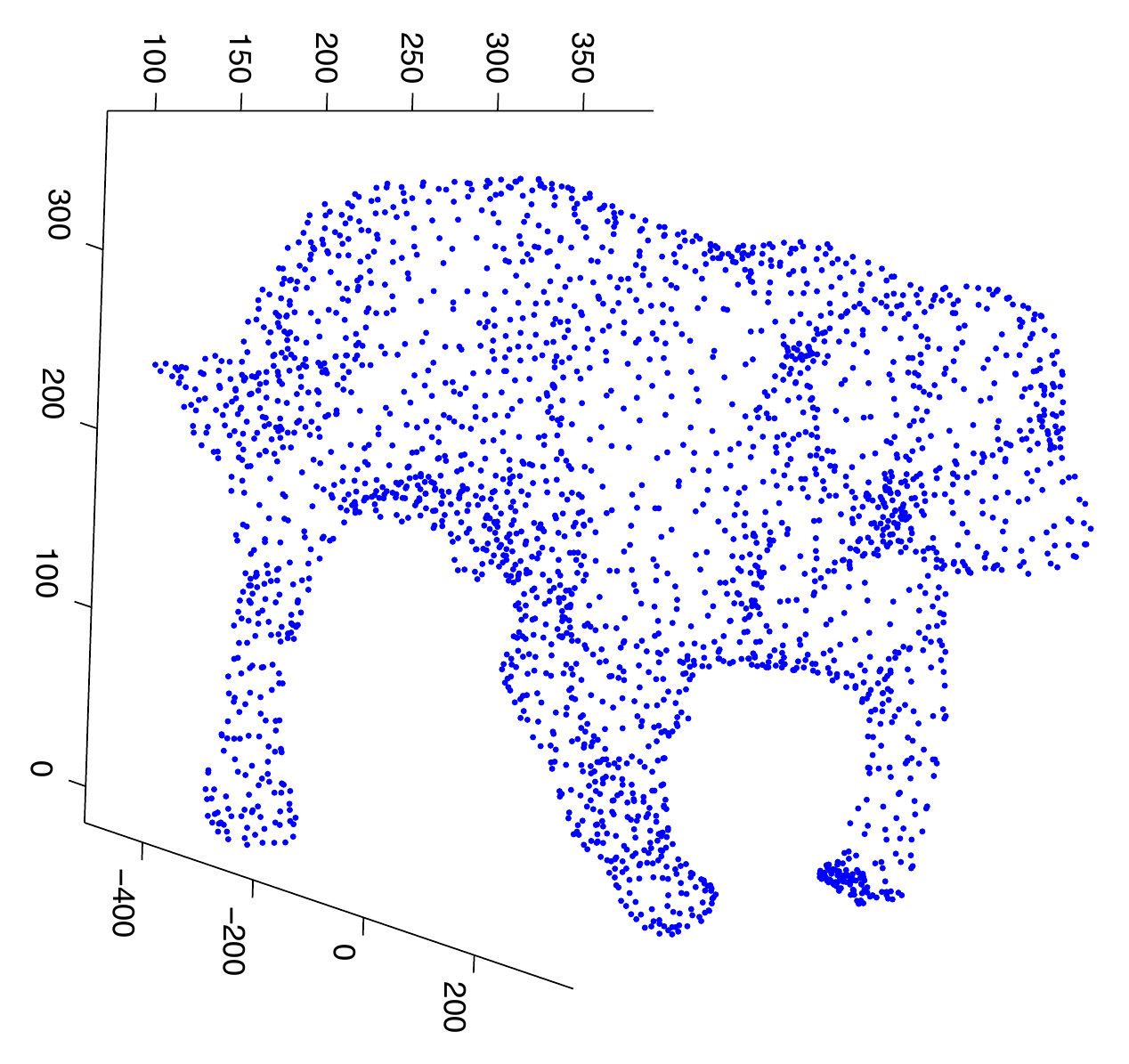}   \label{depth0}}
            \subfigure[Depth 1]{ \includegraphics[width=4cm]{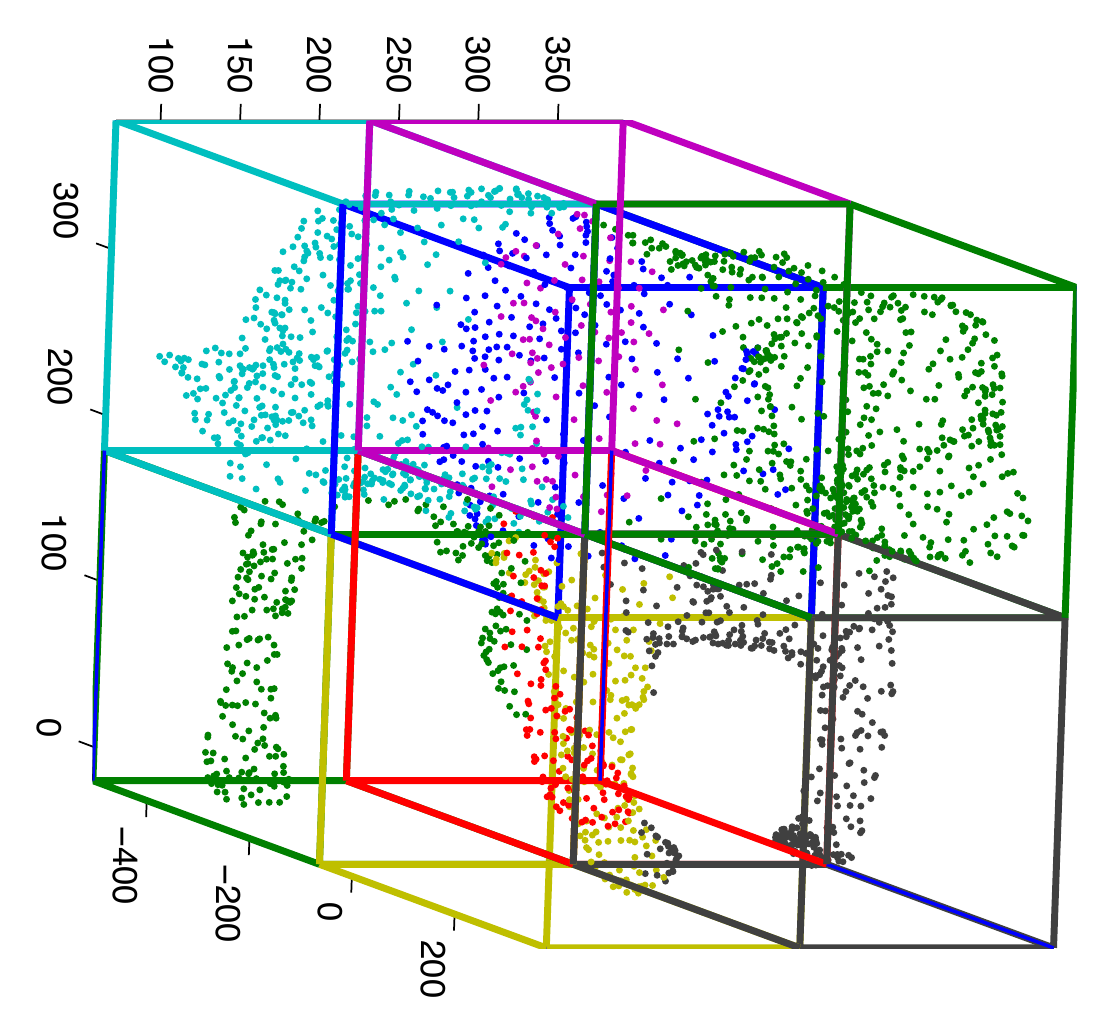} \label{depth1}} 
            \subfigure[Depth 2]{ \includegraphics[width=4cm]{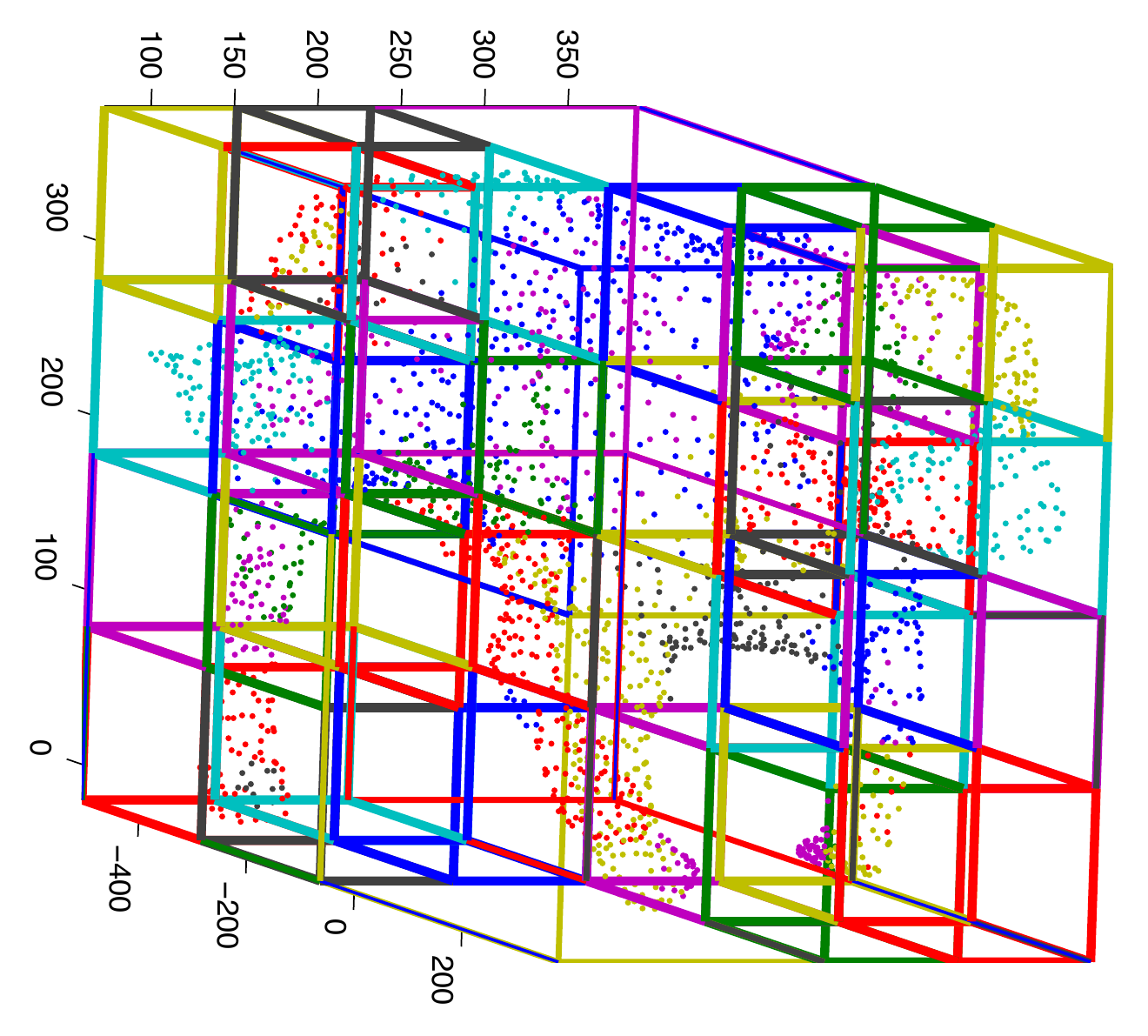}   \label{depth2}}
                    \caption{ Octree decomposition of a 3D model for two different depth levels. The points belonging to each voxel are represented by the same color.  } 
        \label{fig:octree_decomposition}
\vspace{-0.1cm}
\end{figure*}

 \section{Related work}
\label{sec:related_work}
 
The direct compression of 3D point cloud sequences has been largely overlooked so far in the literature. A few works have been proposed to compress static 3D point clouds. Some examples include the 2D wavelet transform based scheme of  \cite{Ochotta_2004}, and the subdivision of the point cloud space in different resolution layers using a kd-tree structure \cite{Devillers_2000}.  An efficient binary description of the spatial point cloud distribution is performed through a decomposition of the 3D space using octree data structures. The octree decomposition,   in contrast to the mesh construction, is quite simple to obtain.  It is the basic idea behind the geometry compression algorithms of \cite{schnabel_2006}, \cite{Huang2008}. The octree structure  is also adopted  in \cite{Zhang2014}, to compress point cloud attributes.    The authors  construct a graph for each branch of leaves at certain levels of the octree. The graph transform, which is equivalent to the Karhunen-Lo\`{e}ve transform, is then applied to decorrelate the color attributes that are treated as signals on the graph. The proposed algorithm has been shown  to remove the spatial redundancy for compression of the 3D point cloud attributes, with significant improvement over traditional methods.  However, all the above methods  are  designed mainly for static point clouds. In order to apply them to point cloud sequences, we need to consider each frame of the sequence independently, which is clearly suboptimal. 
 
 Temporal and spatial redundancy of point cloud sequences has been recently exploited in \cite{Kammerl2012}. The authors compress the geometry by comparing the octree data structure of consecutive point clouds and encoding their structural difference. The proposed compression framework can handle general point cloud streams of arbitrary and varying size, with unknown correspondences. It enables detection and differential encoding of spatial changes within temporarily adjacent octree structures by modifying the octree data structure, without though computing the exact motion of the voxels.     Motion estimation in point clouds sequences can be quite challenging due to the fact that point-to-point correspondences between consecutive frames are not known. While there exists a huge amount of works in the literature that study the problem of motion estimation in video compression, these methods cannot be extended easily to graph settings. In classical video coding schemes, motion in 3-D space is mainly considered as a set of  displacements in the regular image plane.   Pixel-based methods \cite{Irani_1999}, such as block matching algorithms,    or  optical and scene flow algorithms, are designed for regular grids. Their generalization to the non-Euclidean, irregular graph domain though is not straightforward.  
 Feature-based methods   \cite{Torr_1999}, such as interest point detection, have also been widely used for motion estimation in video compression. These features usually correspond to key points of images such as corners or sharp edges \cite{Lowe_2004, Bay_2006, Extremal_2002}. With an appropriate definition of features on graphs, these methods can be extended to graphs.  To the best of our knowledge though, so far they are not adapted to estimate the motion  on graphs, nor on point clouds. Someone could also apply classical 3D descriptors such as \cite{Scovanner_2007, Tombari_2010, Zaharescu_2009, Tombari_2011, Chen_2007, Sipiran_2010} to define 3D features. However, these type of descriptors assume that the point cloud represents a surface, which is not necessarily the case of a graph. Moreover, most of them require the computation of the normals which can be complex to obtain in real-time scenarios. An overview of classical 3D descriptors can be found in \cite{Alexandre_2012}. 
 
For the sake of completeness, we should mention that  3D point clouds are often converted into polygonal meshes, which can be compressed with a large body of existing methods.   In particular, there does exist literature for compressing dynamic 3D meshes with either fixed connectivity and known correspondences (e.g., \cite{Peng_2005, Rossignac_1999, Alexa_2000, Vasa_2010, Nguyen_2014}) or varying connectivity (e.g., \cite{Han_2007, Gupta_2003}).  A different type of approach consists of the video based methods. The irregular 3D structure of the meshes is parametrized into a rectangular 2D domain, obtaining the so called geometry images \cite{Gu_2002} in the case of a single mesh and geometry videos \cite{Briceno2003} in the case of 3D mesh sequences. The mapping of the 3D mesh surface onto a 2D array, which can be done either by using only the 3D geometry information or both the geometry and the texture information \cite{Habe_2004},  allows conventional video compression to be applied to the projected 2D videos. Within the same line of work, emphasis has been given to extending these types of algorithms to handling sequences of meshes with different numbers of vertices and exploiting temporal correlation between them.   An example is the recent work in  \cite{Hou_2015},  which proposes a framework for compressing 3D human motion oriented geometry videos by constructing key frames that are able to reconstruct the whole motion sequence.   Comparing to the mesh-based compression algorithms, the advantage is that the mesh connectivity information does not need to be sent, and the complexity is reduced by performing the operations from the 3D to the 2D space. All the above mentioned works however require the conversion process of the point cloud into a mesh, which is usually computationally expensive, and it cannot be easily applied in real-time applications.  Finally, marching cubes  algorithm \cite{Lorensen_1987} can be used to extract a polygonal mesh in a fast way, but it requires a ``filled" volume.

\section{Structural representation of 3D point clouds}
\label{sec:representation_with_graphs}
3D point clouds usually have little explicit spatial structure.  Someone can however organize the 3D space by converting the point cloud into an octree data structure \cite{Loop_2013}, \cite{Huang2008}, \cite{schnabel_2006}. In what follows, we  recall the octree construction process, and introduce graphs as a tool for capturing the structure of the leaf nodes of the octree.  

\subsection{Octree representation of 3D point clouds}

 An octree is a tree structure with a predefined depth, where every branch node represents a certain cube volume in the 3D space, which is called a voxel. 
A voxel containing a sample from the 3D point cloud is said to be occupied. Initially, the 3D space is  hierarchically  partitioned into voxels whose  total number  depends on the number of subdivisions, i.e., the depth of the resulting tree structure.   For a given depth, an octree is constructed by traversing the tree structure  in depth-first order. Starting from the root, each node can generate eight children  voxels.  At the maximum depth of the tree, all the points are mapped to   leaf voxels. 
An example of the voxalization of a 3D model for different depth levels, or equivalently for different quantization stepsizes, is shown in Fig.~\ref{fig:octree_decomposition}. 

In contrast to polygonal mesh representations, the octree structure is easy to obtain  and effective in real-time applications.  Thanks to the different depths of the tree,  it allows a multiresolution representation of the data that leads to efficient data processing  in many applications.    In particular, this multiresolution representation permits a progressive compression of the 3D positions of the data, which is lossless within each representation level \cite{Kammerl2012}.  

\subsection{Graph-based representation of 3D point clouds}
Although the overall voxel set lies on a regular grid, the set of occupied voxels is non-uniformly distributed in space, as most  of the leaf voxels are unoccupied.  In order to represent the irregular structure formed by the occupied voxels, we use a graph-based representation. Graph-based representations are flexible and well adapted to data that live on an irregular domain \cite{Shuman2013}. In particular, we represent the set of occupied voxels of the  octree using a weighted and undirected graph $\mathcal{G}=(\V,\E,W)$,  where $\V$ and $\E$ represent the vertex and edge sets of  $\mathcal{G}$.  
Each of the $N$ nodes in $\V$ corresponds to an occupied voxel, while each edge in $\E$ connects neighboring, occupied voxels. We define the connectivity of the graph based on the $K$- nearest neighbors ($K$-NN graph), which is widely used in the literature. We usually set $K$ to 26 as it corresponds to the maximum number of neighbors for a node that have a maximum distance of one step along any axis of the 3D space. Two vertices are thus considered to be  neighbors  if they are among  the 26 nearest neighbors in the voxel grid. 
The matrix $W$ is a matrix of positive edge weights, with $W(i,j)$ denoting the weight of an edge connecting vertices $i$ and $j$. This weight captures the connectivity pattern of nearby occupied voxels and are chosen to be inversely proportional to the distances between voxels.

\subsection{Graph Fourier transform of 3D point cloud attributes}
After the graph $\mathcal{G}=(\V,\E,W)$ is constructed, we consider the attributes of the 3D point cloud --- the 3D coordinates $p = [ x,y,z]^T\in\mathbb{R}^{3\times N}$ and the color components $c=[r,g,b]^T\in\mathbb{R}^{3\times N}$ --- as signals that reside on the vertices of the graph $\G$. A spectral representation of these signals can be obtained with the help of the Graph Fourier Transform (GFT). The GFT is defined through the graph Laplacian operator $\L=D- W$,  where $D$ is the diagonal degree matrix whose $i^{th}$ diagonal element is equal to the sum of the weights of all the edges incident to vertex $i$ \cite{Chung97}. The graph Laplacian is a real symmetric matrix that has a complete set of orthonormal eigenvectors with corresponding nonnegative eigenvalues. We here denote its eigenvectors  by $\chi=[\chi^{0},\chi^{1},...,\chi^{N-1}]$, and the spectrum of eigenvalues by 
$\Lambda:=\Bigl\{0=\lambda_{0}\le\lambda_{1}\le\lambda_{2}\le...\le\lambda_{(N-1)}\Bigr\}$, where $N$ is the number of vertices of the graph. The multiplicity of the smaller eigenvalue 
indicates the number of connected components of the graph. The GFT of any graph signal $f\in\Rbb^N$ is then defined as 
\begin{equation*}
F_f(\lambda_{\ell}) := <f,\chi_{\ell}> = \sum_{n = 1}^N f(n)\chi_{\ell}^*(n),  
\end{equation*} 
while the inverse graph Fourier transform is given by 
\begin{equation*}
f(n) = \sum_{\ell = 0}^{N-1}F_f(\lambda_{\ell}) \chi_{\ell}(n).
\end{equation*} 
The GFT is useful to have an effective representation of the data. Furthermore, it has been shown to be optimum for decorrelating a signal following the  Gaussian Markov Random Field model with precision matrix $\L$ \cite{Zhang_2013}. The GFT will be used later to define features in point cloud frames, and for effective coding of the data on the graph.

\section{Motion estimation in 3D point cloud sequences}
\label{sec:Motion_estimation_compensation}

As the frames have irregular structures, we use a feature-based matching approach to find correspondences. We use the  graph information  and the signals residing on its vertices to define feature descriptors on each vertex. We first define  simple octant indicator functions to capture the signal values in different orientations.  We then characterize the local topological context of each of the point cloud signals in each of these orientations,   by using spectral graph wavelets (SGW)  computed on the color and geometry signals  at different resolutions \cite{Hammond2010}.   Our feature descriptors, which consist of the wavelet coefficients of each of the signals placed in the corresponding vertex, are 
then used to compute point-to-point correspondences between graphs of different frames.  
We select a subset of best matching nodes to define a sparse set of motion vectors that describe the temporal correlation in the sequence. A dense motion field is eventually interpolated from the sparse set of motion vectors to obtain a complete mapping between two frames. The overall procedure is  detailed below. 

\subsection{Multi-resolution features  on  graphs}
 We define features in each node by computing the variation of the signal values, i.e., geometry and color components, in different parts of its neighborhood.  For each node $i$  belonging to the vertex set $\V$ of a  graph $\mathcal{G}$, i.e., $i\in\V$, we first define the  octant indicator function  $o_{k,i}\in\Rbb^N, \forall ~k = [1,2,...,8]$, given as follows for the first octant 
 \begin{equation*}
\nonumber o_{1,i}(j)  =  \textbf{1}_{\{x(j)\ge x(i), y(j)\ge y(i), z(j)\ge z(i)\} }(j),
 \end{equation*}
where $\textbf{1}_{\{\cdot\}}(j)$ is the indicator function on  $j\in\V$, evaluated in a set ${\{\cdot\}}$  of voxels given by specific  3D coordinates. The first octant indicator function  is thus nonzero only in the entries corresponding to the voxels whose  3D position coordinates are bigger than the ones of node $i$.    We consider all possible combinations of coordinates, which results in a total of $2^3$ indicator functions for the eight octants around $i$.  These functions result in a clustering of the nodes of the graph, providing a notion of orientation of each node in the 3D space with respect to $i$, which is clearly provided by the voxel grid. 
 
 We then compute features based on both geometry and color information, by treating their values independently in each orientation. In particular, for each node $i\in\V$ and each geometry and color component $f\in\Rbb^N$, where $f\in \{x,y,z,r,g,b\}$, we compute the spectral graph wavelet coefficients by considering independently the values of $f$ in each orientation $k$ with respect to node $i$ such that
\begin{equation}
\phi_{i,s,o_{k,i},f} = <f\cdot o_{k,i},\psi_{s,i}>, 
\label{eq:wavelet_feature}
\end{equation}
where  $k \in \{1,2,...,8\}$, and $ s \in \mathcal{S} = \{s_1,...,s_{max}\}$, is a set of discrete scales. The function $\psi_{s,i}$ represents the spectral graph wavelet of scale $s$ placed at that particular node $i$, and $\cdot$ denotes the pointwise product. 
We recall that the spectral graph wavelets \cite{Hammond2010}  are operator-valued functions of the graph Laplacian defined as 
\begin{equation*}
\psi_{s,i} = T^s_g\delta_i = \sum_{\ell = 0}^{N-1}g(s\lambda_{\ell})\chi_{\ell}^{*}(i)\chi_{\ell}.
\end{equation*}
The graph wavelets are determined by the choice of a generating kernel $g$, which acts as a band-pass filter in the spectral domain, and a scaling kernel $h$ that acts as a lowpass filter. The scaling is defined in the spectral domain, i.e., the wavelet operator at scale $s$ is given by $T^s_g = g(s\mathcal{L})$. Spectral graph wavelets are finally realized through localizing these operators via the impulse $\delta$ on a single vertex $i$. The scaling function $h$, which is analogous to the lowpass scaling function in classical wavelet analysis, captures the low frequency content of the signals \cite{Hammond2010}. It thus acts as a low pass filter and helps ensure stable recovery of the graph signal from the wavelet coefficients. The application of these wavelets to  signals living on the graph results in a multi-scale descriptor for each node. The descriptor  characterizes the local topological context of the signals in the different  neighborhoods defined by the octant indicator functions.    We define the feature vector $\phi_i$ at node $i$ as the concatenation of the coefficients computed in (\ref{eq:wavelet_feature}) with wavelets at different scales, including the features obtained from the scaling function, i.e., $\phi_i = \{\phi_{i,s,o_{k,i},f}\}\in\mathbb{R}^{8\times6\times(
|\mathcal{S}|+1)}$. These features can be efficiently computed by approximating  the spectral graph wavelets with Chebyshev polynomials, as described in \cite{Hammond2010}. Given this approximation, the wavelet coefficients at each scale can then be computed as a polynomial of $\L$ applied to the graph signal $f$.  The latter can be performed in a way that accesses $\L$ only through iterative matrix-vector multiplications. The polynomial approximation can be  particularly efficient when the graph is sparse, which is indeed the case of our $K$-NN graph. Moreover, this approximation avoids the need to compute the complete spectrum of the graph Laplacian matrix.  Thus, the computational cost of the features can be substantially reduced.   

Finally, we note that spectral features have  recently started to gain attention in the computer vision and shape analysis community.  The heat kernel signatures \cite{Sun_2009}, their scale-invariant version \cite{Bronstein_2010}, the wave kernel signatures \cite{Aubry_2011}, the optimized spectral descriptors of \cite{Litman_2014}, have already been used in 3D shape processing with applications in graph matching \cite{Hu_2014_CVPR} or in mesh segmentation and surface alignment problems \cite{KimCS13}. These features have been shown to be stable under small perturbations of the edge nodes of the graph. In all these works though, the descriptors are defined based only on the graph structure, and  the information about  attributes of the nodes such as color and 3D positions, if any,  is assumed to be introduced in the weights of the graph.    The performance of these descriptors depends   on the quality of the defined graph. In contrast to this line of works, we define features by considering attributes as signals that reside on the vertices of a graph and characterize each vertex by computing the local evolution of these signals at different scales.  Furthermore, this approach gives us the flexility to consider the signal values in different orientations as discussed above, and makes the descriptor of each node more informative.

\subsection{Finding correspondences on dynamic graphs}
We translate the problem of finding correspondences in two consecutive point clouds or frames of the sequence into finding correspondences on the vertices of their representative graphs.   For the rest of this paper, we denote the sequence of frames as $\mathcal{I} = \{\mathcal{I}_1,~\mathcal{I}_2,~...,~\mathcal{I}_{max}\}$ and the set of graphs corresponding to each frame as $\mathcal{G} = \{\mathcal{G}_1,~\mathcal{G}_2,~...,~\mathcal{G}_{max}\}$. For two consecutive frames of the sequence, $\mathcal{I}_t,~\mathcal{I}_{t+1}$, referred also as reference and target frame respectively, our goal is to find correspondences between the vertices of their representative graphs $\mathcal{G}_t$ and $\mathcal{G}_{t+1}$. The number of vertices can differ between the graphs and is denoted as $N_t$ and $N_{t+1}$ respectively.  

 Given  two graphs $\mathcal{G}_t$, $\mathcal{G}_{t+1}$,  and their respective vertex sets $\V_t, ~\V_{t+1}$,  we use the  features defined in the previous subsection to measure the similarity between vertices. We compute the matching score between two nodes $m\in\V_t, n\in\V_{t+1}$ as the Mahalanobis distance between the corresponding feature vectors, i.e.,
\begin{equation}
\small{
\sigma(m,n) = (\phi_m-\phi_n)^T P (\phi_m-\phi_n), \quad \forall m\in\V_{t}, ~n\in\V_{t+1},
}
\label{eq:mah_distance}
\end{equation}
where $P$ is a matrix that characterize the relationships between the geometry and the color feature components, which are measured in different units, as well as the contribution of each of the wavelet scales in the matching performance. We learn the  positive definite matrix $P$ by estimating the sample inverse  covariance matrix from a set of training features that are known to be in correspondence. If $m\in\mathcal{V}_t$ corresponds to $n\in\mathcal{V}_{t+1}$, this models $\phi_m$ as a Gaussian random vector with mean $\phi_n$ and covariance $P^{-1}$, while if $m$ does not correspond to $n$, this models $\phi_m$ as coming from a very flat (essentially uniform) distribution. Hence the matching score $\sigma(m,n)$ can be considered  a log likelihood ratio for testing the hypothesis that $m$ corresponds to $n$.

For each node in $\G_{t+1}$, we use the matching score to  define  its best matching node in $\G_t$. In particular, for each  $n\in\V_{t+1}$, we define as its best match in $\V_{t}$, the node $m_n$ with the minimum Mahalanobis distance, i.e.,
\begin{equation*}
\nonumber m_n = \argmin_{m\in\V_t} \sigma(m,n).
\end{equation*}   
 From the global set of correspondences computed for all the nodes of $\V_{t+1}$, we select   a sparse set of significant matches.   The objective of this selection is to take into consideration only accurate matches and ignore others since inaccurate correspondences are possible with our spectral descriptors in the case of large displacements. We also want to avoid matching  points in $\mathcal{I}_{t+1}$ that do not have any true correspondence in the preceding frame $\mathcal{I}_t$.  The sparse set of matching nodes will later be used for  interpolating the motion across all the nodes of the graph. For that reason, we need to ensure that we keep correspondences in all areas of the 3D space, and these correspondences should be spatially discriminative. We cluster the vertices of $\mathcal{G}_{t+1}$ into different regions and we keep only one correspondence, i.e., one  representative vertex, per region. Clustering is performed by applying $K$-means in the 3D coordinates of the nodes of the target frame, where $K$ is usually  set to be equal to the target number of significant matches. In order to avoid inaccurate matches, a representative vertex per cluster is included in the sparse set only  if its best score is smaller than a predefined threshold. This  procedure results in detecting a sparse set of vertices $n$ in $\V_{t+1}$, denoted $\V_{t+1}^S \subset \V_{t+1}$, and the set of their correspondences $m_n$ in $\V_t$, $\V_{t}^S \subset \V_{t}$. Moreover, our sparse set of matching points tend to be accurate correspondences that are well distributed spatially.

\subsection{Computation of the motion vectors}

We now describe how we generate a dense motion field from the sparse set of matching nodes $(m_n,n)\in\V_{t}^S\times\V_{t+1}^S $.  Our implicit assumption is that vertices that are close in terms of 3D positions, which means that they are close neighbors in the underlying graph, undergo a similar motion.  We thus use the structure of the graph in order to interpolate the motion field, which is assumed to be smooth on the graph. 

In more detail, our goal is to estimate the dense motion field $v_t = [v_t(m)]$, for all $m\in\mathcal{G}_t$, using the correspondences $(m_n,n)\in\V_{t}^S\times\V_{t+1}^S $.  To determine $v_t(m)$ for $m=m_n\in\mathcal{V}_t^S$, we use the vector between the pair of matching points $(m_n,n)$,
\begin{equation}
v_t(m_n|n) \stackrel{\Delta}{=} p_{t+1}(n)- p_{t}(m_n).
\label{eqn:motion_vector}
\end{equation}
Here we recall that $p_t$ and $p_{t+1}$ are the 3D positions of the vertices  of $\mathcal{G}_t$ and $\mathcal{G}_{t+1}$, respectively. To determine $v_t(m)$ for  $m\not\in\mathcal{V}_t^S$, we consider the motion field $v_t$, like $p_t$, to be a vector-valued signal that lives on the vertices of  $\mathcal{G}_t$. Then we smoothly interpolate the sparse set of motion vectors (\ref{eqn:motion_vector}).  The interpolation is performed by treating each component independently. Given the motion values on some of the vertices, we pose interpolation as a regularization problem that estimates the motion values on the rest of the vertices by requiring the motion signal to vary smoothly across vertices that are connected by an edge in the graph.   Moreover, we allow some smoothing on the known entries.   The reason for that is that the proposed matching scheme does not necessarily guarantee that the sparse set of correspondences, and the estimated motion vectors associated with them,  are correct. To limit the effect of motion estimation inaccuracies,  for each matching pair $(m_n,n)\in\V_{t}^S\times\V_{t+1}^S $, we model the matching score in the local neighborhood of $m_n\in\V_t^S$ with a smooth signal approximation. Specifically, for each $n\in\V_{t+1}^S$, we extend the definition (\ref{eqn:motion_vector}) to all $m\in\V_t$, i.e.,
\begin{equation*}
v_t(m|n) = p_{t+1}(n)- p_{t}(m).
\end{equation*}
Then, for each node that belongs to the two-hop neighborhood of $m_n$ i.e., $m\in\mathcal{N}_{m_n}^2$, we express $\sigma(m,n)$ as a function of the geometric distance of $p_t(m)$ from $p_t(m_n)$, using a  second-order Taylor series expansion  around $p_t(m)$.  That is,
\begin{align}
\nonumber  &\sigma(m,n)  \approx \sigma(m_n,n) \\
\nonumber &+ (p_t(m)- p_t(m_n))^T {M_n^{-1}}(p_t(m)- p_t(m_n))\\
\nonumber & = \sigma(m_n,n) \\
&+ (v_t(m|n) - v_t(m_n|n))^T {M_n^{-1}}(v_t(m|n) - v_t(m_n|n)). \label{eq:taylor_approx}
\end{align}
For each $n\in\mathcal{V}_{t+1}^S$, we take $ \sigma(m,n)$ to be a discrete sampled version of a continuous function $ \sigma(v,n) $ where the second order Taylor approximation is
\begin{equation*}
\sigma(v,n)  \approx \sigma(m_n,n) + (v - v_t(m_n|n))^T {M_n^{-1}}(v - v_t(m_n|n) )).
\end{equation*}
Thus for each $n\in\V_{t+1}^S$, we assume that the matching score with respect to  nodes that are in the neighborhood of its best match $m_n\in\V_{t}^S$ can be well modeled by a quadratic approximation function. 
We  estimate $M_n$ of this quadratic approximation as the normalized covariance matrix of the 3D offsets, 
\begin{equation*}
\small{
 M_n = \frac{1}{|\mathcal{N}_{m_n}^2|} \sum_{m\in\mathcal{N}_{m_n}^2 } \frac{ (p_t(m)- p_t(m_n))(p_t(m)- p_t(m_n))^T }{\sigma(m,n) - \sigma(m_n,n)}.}
\end{equation*}
This is motivated by the fact that if $\sigma(m,n) - \sigma(m_n,n) =(v_t(m) - v_t(m_n|n))^T {M_n^{-1}}(v_t(m) - v_t(m_n|n) )$, then
\[
u = \frac{v_t(m) - v_t(m_n|n)}{\sqrt{\sigma(m,n) - \sigma(m_n,n)}}
\]
satisfies $1 = u^TM_n^{-1}u$. Hence, $u$ lies in an ellipsoid whose second moment is proportional to $M_n$. Though there are other ways for computing $M_n$ in (\ref{eq:taylor_approx}), this moment-matching method is fast while guaranteeing that $M_n$ is positive semi-definite.  Next, we use the covariance matrices of the 3D offsets to define a diagonal matrix $Q\in \Rbb^{3N_t\times 3N_t}$, such that 
\[
Q = \begin{bmatrix}
       M_1^{-1} & \cdots & \mathbf{0}_{3\times 3} \\[0.3em]
        \vdots & \ddots &  \vdots         \\[0.3em]
       \mathbf{0}_{3\times 3}         & \cdots  & M_{N_t}^{-1}
     \end{bmatrix},
\]
where $M_{m} = M_n$ if $m=m_n$ for some $n\in \V_{t+1}^S$, and $M_{m} = \mathbf{0}_{3\times 3}$  otherwise. 
The matrix $Q$ captures the second order Taylor approximation of the total match score as a function of the   motion vectors in the neighborhoods of nodes in $\V_t^S$ and is used to regularize the motion vectors of the known entries in $v_t$ as shown next.  
     
Finally, we interpolate the dense set of motion vectors  ${v_t}^{*}$ by taking into account the covariance of the motion vectors in the  neighborhoods around the points  that belongs to the sparse set $\mathcal{V}_t^S$ and  imposing smoothness of  the  dense motion vectors on the graph  
\begin{equation}
{v_t}^{*} =  \argmin_{v\in\mathbb{R}^{3N_t}} (v-v_t)^T Q (v-v_t) + \mu \sum_{i = 1}^3 (S_i v)^T \mathcal{L}_t(S_i v),
\label{eq:smoothing_eq}
\end{equation}
where $\left\{S_i\right\}_{i=1,2,3}$  is a selection matrix for each of the 3D components respectively, and $\mathcal{L}_t$ is the Laplacian matrix of the graph $\mathcal{G}_t$.  $v_t= [v_t(1), v_t(2),\cdots,v_t(N_t)]^T\in\mathbb{R}^{3N_t}$ is the concatenation of the initial motion vectors, with $v_t(m) = \mathbf{0}_{3\times1}$, if $m\notin\V_t^S$. We note that the optimization problem consists of  a fitting term that penalizes the excess matching score on the sparse set of matching nodes, and of a regularization term that  imposes smoothness of the motion vectors in each of the position components independently. The tradeoff between the two terms is defined by the constant $\mu$. A small $\mu$ promotes  a solution that is closed to $v_t$, while a big $\mu$  favors a solution that is very smooth.   Similar regularization techniques, which are based on the notion of smoothness of the  graph Laplacian, have been widely used in  the  semi-supervised learning literature \cite{Zhu03, zhou04learningLocalGlobal}. 
The corresponding optimization problem is convex and it has a closed form solution given by 
\begin{equation}
v_t^{*} = \big(Q + \mu \sum_{i = 1}^3 S_i^T \mathcal{L}_t  S_i \big)^{-1} Q v_t,
\label{eq:closed_form}
\end{equation}
which can be computed  iteratively using MINRES-QLP \cite{Choi2014} in large systems. With a slight abuse of notation, we will from now on denote as $v^*_t$ the reshaped motion vectors of dimensionality $3\times N_t$, where each row represents the motion in one of the three coordinates. Finally, $v^*_t(m) \in \Rbb^3$ denotes  the 3D motion  vector of node $m\in\V_t$.

\section{Compression of 3D point cloud sequences}
\label{sec:coding_of_3D_sequences}
 We describe now how the above motion estimation can be used to reduce temporal redundancy in compression of 3D point cloud sequences, by describing in detail each block of Fig.~\ref{fig:figure_intro}.  We first  code  the motion vectors by transforming them to the graph Fourier domain. Coding of the 3D positions is then performed by comparing  the structural difference between the target frame ($\mathcal{I}_{t+1}$) and the motion compensated reference frame ($\mathcal{I}_{t,mc}$). Temporal redundancy in color compression is finally exploited by encoding the difference between the target frame  and the color prediction obtained from motion compensation.  
 
\subsection{Coding of motion vectors} 
\label{subsec:motion_vector_coding}
\begin{figure}[]
      \centering
         
          { \includegraphics[width=9cm]{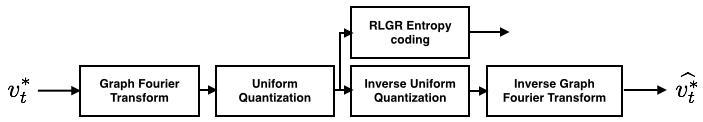} \label{compression_motion_vectors}} 
  
           \caption{Schematic overview of the motion vector coding scheme. The motion vectors $v^*_t$ between two consecutive frames of the sequence are transformed in the graph Fourier domain, quantized uniformly, and sent to the decoder. The decoder performs the reverse procedure to obtain $\widehat{v_t^*}$. } 
        \label{fig:compression_mv}
\vspace{-0.1cm}
\end{figure}

We recall that, for each pair of two consecutive frames $\mathcal{I}_t, ~ \mathcal{I}_{t+1}$, the sparse set of motion vectors is initially smoothed at the encoder side. The estimated dense motion field is then transmitted to the decoder. One could transmit only the motion vectors on the sparse set of matching points and solve the interpolation problem (\ref{eq:closed_form}) at the decoder. That would however increase the complexity of the decoder. We exploit the fact that the graph Fourier transform is suitable for compressing smooth signals \cite{Zhu_2012},  \cite{Zhang_2013}, by coding the motion vectors  in the graph Fourier domain.   In particular, since the motion $v^*_t$ is estimated in each of the nodes of the graph $\mathcal{G}_t$, we use the eigenvectors $\chi_t=[\chi^{0}_t,\chi^{1}_t,...,\chi^{N_t-1}_t]$ of the graph Laplacian operator  corresponding to the graph $\mathcal{G}_t$ of the reference frame,   to transform the motion in each of the directions separately such as
\begin{equation*}
F_{v^{*}_t}(\lambda_{\ell}) = <{v}^{*}_t, \chi^{\ell}_t>, \quad \forall \ell = 0,1,...,N_t-1.
\end{equation*}  
The transformed coefficients are uniformly quantized as $round(\frac{F_{v^{*}_t}}{\Delta})$, where $\Delta$ is the quantization stepsize that is constant across all the coefficients, and $round$ refers to the rounding operation. The choice of the quantization stepsize will be discussed in the  experimental section. The quantized coefficients are then entropy coded independently with the adaptive run-length / golomb-rice (RLGR) entropy coder  \cite{Malvar_2006} and sent to the decoder. The decoder performs  the reverse procedure to obtain the decoded motion vectors $\widehat{v_t^*}$.  Note that given that the decoder already knows the 3D positions of the reference frame, it can recover the $K$-NN graph. Thus,  the connectivity of the graph does not have to be sent to the decoder.  A block diagram of the encoder and the decoder is shown in Fig.~\ref{fig:compression_mv}. 

\subsection{Motion compensated differential coding of 3D geometries}
\label{subsec:geometry_coding}
From the current frame $\mathcal{I}_t$ and its quantized motion vectors $\widehat{v_t^*}$, both of which are signals on $\mathcal{G}_t$, it should be possible to predict the 3D positions of the points in the target frame $\mathcal{I}_{t+1}$, which is a signal on $\mathcal{G}_{t+1}$.
Since the two graphs are of different sizes, a vector space  prediction of $\mathcal{I}_{t+1}$ from $\mathcal{I}_t$ is not possible. One can however warp $\mathcal{I}_t$ to $\mathcal{I}_{t+1}$ in order to obtain a warped frame $\mathcal{I}_{t,mc}$ that is close to $\mathcal{I}_{t+1}$ in some sense. Given that the 3D positions $p_t$  and the decoded motion vectors $\widehat{v_t^*}$ of $\mathcal{I}_t$ are known to both the encoder and the decoder,   the position of  node $m$ in the warped frame $\mathcal{I}_{t,mc}$  can be  estimated on both sides as 
\begin{equation}
p_{t,{mc}} (m)= p_t(m) + \widehat{{v}^*_t}(m), \quad \forall m\in \V_t.
\label{eqn:mc_geometry}
\end{equation}
Note that the warped frame $\mathcal{I}_{t,mc}$ remains a signal on the graph $\mathcal{G}_t$. 

Given the warped frame $\mathcal{I}_{t,mc}$, we use the real-time compression algorithm proposed in \cite{Kammerl2012} to code the structural difference between the 3D positions of $\mathcal{I}_{t+1}$ and $\mathcal{I}_{t,mc}$.   This algorithm essentially codes the set difference between the set of voxels occupied by the points of $\mathcal{I}_{t+1}$ and the set of voxels occupied by the points of $\mathcal{I}_{t,mc}$.  Specifically, we assume that the  point clouds corresponding to $\mathcal{I}_{t,mc}$ and $\mathcal{I}_{t+1}$ have already been  spatially decomposed into octree data structures at a predefined depth. At each level of the tree, the representation of each octree is done by capturing the occupancies of the children of an occupied voxel with a single byte, whose bits are set to one if the corresponding child is occupied and zero otherwise.   Assuming a consistent order of the octants, each octree is then characterized by a bit stream, whose total size in bits is equal to eight times the number of internal nodes of the tree.  

\begin{figure*}[]
      \centering
            \subfigure[ Differential encoding of consecutive frames ]{ \includegraphics[width=7cm]{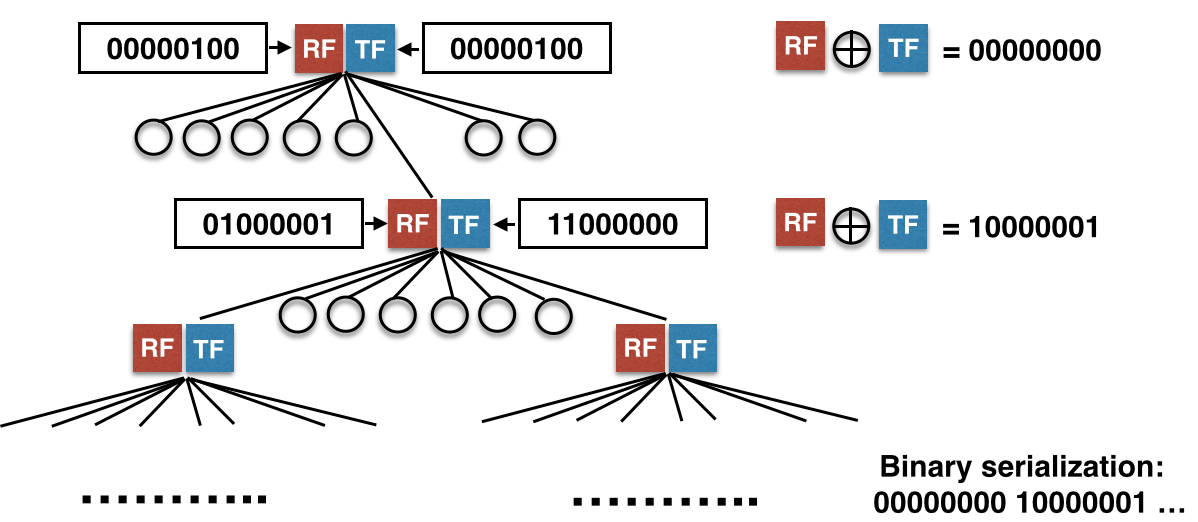}   \label{fig:differetial_encoding}}
            \subfigure[Schematic overview of the compression architecture]  { \includegraphics[width=8cm]{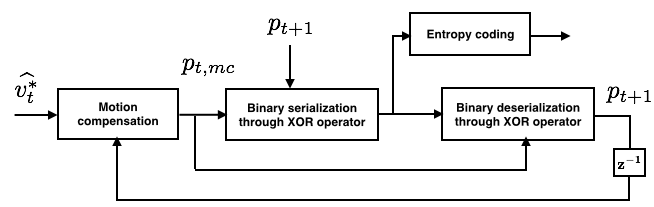} \label{fig:compression_geometry}} 
           \caption{Illustration of the geometry compression of the target frame (TF) based on the motion compensated reference frame (RF).  The differential encoding of the consecutive frames is shown in (a). Structural changes within octree occupied voxels are extracted during the binary serialization process and encoded using the XOR operator.   The bit stream of the XOR operator is sent to the decoder. The figure is inspired by   \cite{Kammerl2012}.  A schematic overview of the overall  3D geometry coding scheme is shown in (b). } 
        \label{fig:compression_geometry_all}
\vspace{-0.1cm}
\end{figure*}

Given that both the encoder and decoder know the occupied voxels of the reference frame $\mathcal{I}_t$ and the motion vectors $\widehat{v_t^*}$, they are able to compute the occupied voxels of the motion compensated reference frame $\mathcal{I}_{t,mc}$.  The encoding of the occupied voxels of the target frame $\mathcal{I}_{t+1}$ is performed by encoding the exclusive-OR (XOR) between the indicator functions for the occupied voxels in frames $\mathcal{I}_{t,mc}$ and $\mathcal{I}_{t+1}$.  This can be implemented by an octree decomposition of the set of voxels that are occupied in $\mathcal{I}_{t,mc}$ but not in $\mathcal{I}_{t+1}$, or vice versa, as illustrated in Fig.~\ref{fig:differetial_encoding}.  Thus, motion compensation is expected to reduce the set difference and hence the number of bits used by the octree decomposition. The decoder can eventually use the motion compensated previous frame and the bits from the octree decomposition to recover exactly the set of occupied voxels (and hence the graph and 3D positions) of the target frame $\mathcal{I}_{t+1}$. A schematic overview of the encoding and decoding architecture is shown in Fig.~\ref{fig:compression_geometry}. A detailed description of the algorithm can be found in the original paper \cite{Kammerl2012}.

\subsection{Motion compensated differential coding of color attributes}
\label{subsec:color_coding}
After coding the 3D positions and the motion vectors, motion compensation is  used to predict the color of the target frame from the motion compensated reference frame. While the 3D positions $p_{t,mc}$ of the points in the warped frame $\mathcal{I}_{t,mc}$ are based on the 3D positions of the previous frame $\mathcal{I}_t$ and the motion field on the graph $\mathcal{G}_t$ according to (\ref{eqn:mc_geometry}), the colors $c_{t,mc}$ of the warped frame $\mathcal{I}_{t,mc}$ can be transferred directly from $\mathcal{I}_t$ according to
\[
c_{t,mc}(m) = c_t(m), \quad \forall m\in \V_t.
\]
Unfortunately, the graphs $\mathcal{G}_{t}$ and $\mathcal{G}_{t+1}$ have different sizes; as a consequence there is no direct correspondence between their nodes. However, since $\mathcal{I}_{t,mc}$ is obtained by warping  $\mathcal{I}_t$ to $\mathcal{I}_{t+1}$, we can use the colors of the points in $\mathcal{I}_{t,mc}$ to predict the colors of nearby points in $\mathcal{I}_{t+1}$.  To be specific, for each $n\in\V_{t+1}$, we compute a predicted color value $\widetilde{c_{t+1}}(n)$  by finding the nearest neighbors  $\mbox{NN}_n$  in terms of the Euclidean distance of the 3D positions $p_{t+1}(n)$ and  $p_{t,mc}$, and attributing to $n$ their average color  i.e.,
\begin{equation*}
\widetilde{c_{t+1}}(n) = \frac{1}{|\mbox{NN}_n|}\sum_{m\in\mbox{NN}_n} c_{t,mc}(m),
\end{equation*}
where the cardinality of $\mbox{NN}_n$, i.e., $|\mbox{NN}_n|$,  is usually set to 3. 

Temporal redundancy in the color information  is then removed  by coding only the residual of the target frame with respect to the color prediction obtained with the above method, i.e., $\Delta c_{t+1} = c_{t+1} - \nonumber\widetilde{c_{t+1}}$. 
 The main blocks of the compression scheme are performed using the recently introduced graph-based compression algorithm of \cite{Zhang2014}. The algorithm is designed for compressing  the 3D color attributes in  static frames and it essentially removes the spatial correlation within each frame by coding each color component in the graph Fourier domain.     The algorithm divides  each octree in  small blocks containing $k\times k\times k$ voxels. In each of these blocks, it constructs a graph and computes the graph Fourier transform as described in Section \ref{sec:representation_with_graphs}. We adapt the algorithm to sequences by applying the graph Fourier transform  to the color residual $\Delta c_{t+1}$, and the residuals in each of the three color components   are encoded separately.  The graph Fourier coefficients are quantized uniformly. 

\begin{figure}[]
      \centering
           { \includegraphics[width=8cm]{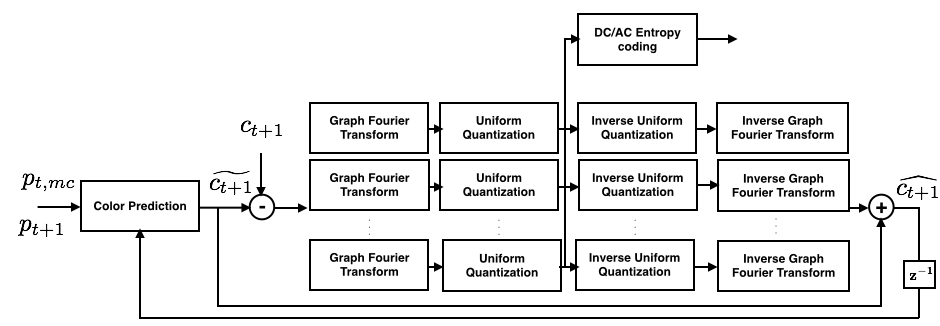} \label{compression_color}} 
           \caption{Schematic overview of the predictive color coding scheme} 
        \label{fig:compression_color}
\vspace{-0.1cm}
\end{figure}

The quantized coefficients are then entropy coded, where the structure of the graph is exploited for better efficiency.  In particular, the coefficients corresponding to the zero eigenvalue of the graph Laplacian represent the DC term as they capture the average of each connected component of the graph. The rest of the coefficients capture higher oscillations of the residual on the graph and they represent the AC term. Due to their different behavior, the DC and the AC coefficients are treated differently during the entropy coding step.   The AC coefficients are assumed to follow a continuous scaled Laplacian distribution, with a diversity parameter inversely proportional to the square root of the corresponding eigenvalue.   
This Laplacian distribution is used by a simple arithmetic encoder to encode the AC components. After encoding a new coefficient, the diversity parameter is then updated as defined in \cite{Zhang2014}.  
The coding of the DC components is first performed by removing the mean of the previously decoded DC terms. The normalized DC coefficient is also assumed to follow a Laplacian distribution with probability function characterized by a diversity parameter that is inversely proportional to the number of connected voxels. 
More details about the color coding scheme are given in \cite{Zhang2014} and a schematic overview  is given in Fig.~\ref{fig:compression_color}. Finally, we recall that while the algorithm was originally used for coding static frames, in this paper we use it for coding the residual of the target frame from the motion compensated reference  frame. The algorithm however remains  a valid choice as the statistics are adapted to the actual signal characteristics. 

\section{Experimental results}
\label{sec:experimental_results}

\begin{figure*}[t]
      \centering
         \subfigure[$ \mathcal{I}_{t} +  \mathcal{I}_{t+1}$]{ \includegraphics[width=2.5cm]{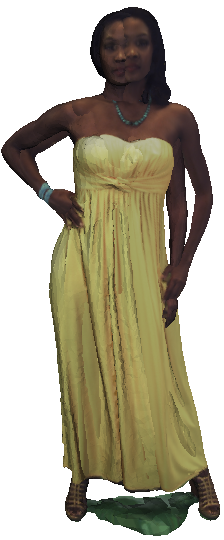}   \label{yellow_dress_example}}
            \subfigure[Correspondence between  $ \mathcal{I}_{t} \mbox{ and } \mathcal{I}_{t+1}$]{ \includegraphics[width=5.6cm]{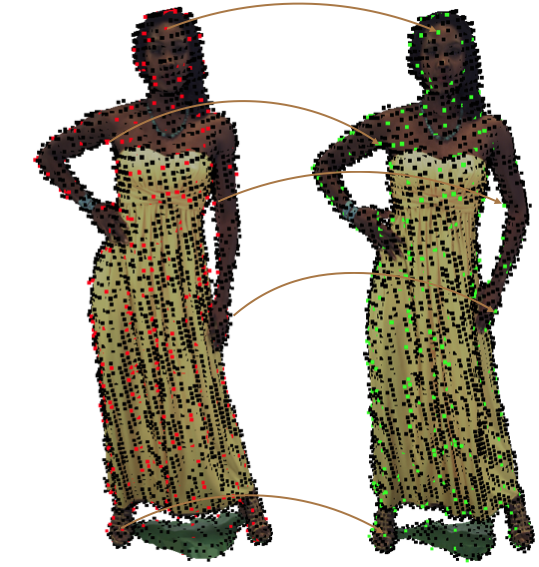} \label{yellow_dress_matching}} 
            \subfigure[$ \mathcal{I}_{t,mc} +  \mathcal{I}_{t+1}$]{ \includegraphics[width=2.7cm]{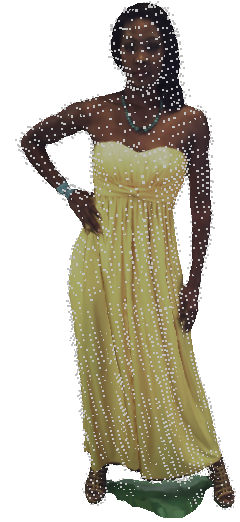}   \label{yellow_dress_motion_compensated}}\\
             \subfigure[$ \mathcal{I}_{t} +  \mathcal{I}_{t+1}$]{ \includegraphics[width=3.8cm]{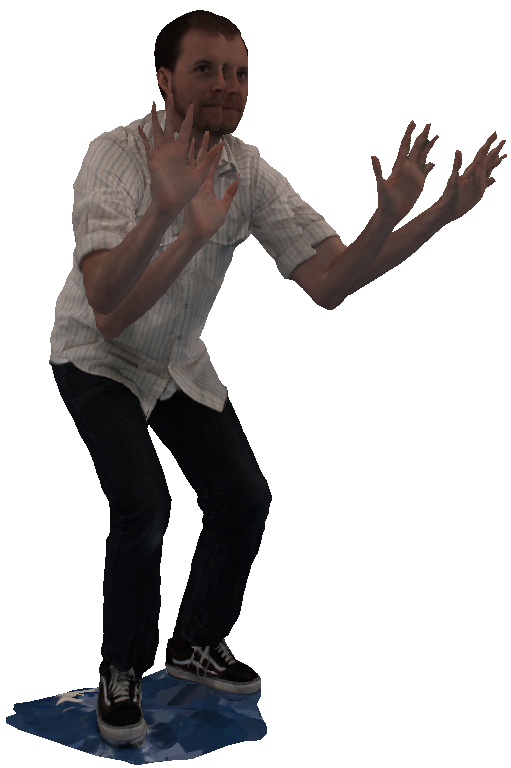}   \label{man_sequence_example}}
            \subfigure[Correspondence between  $ \mathcal{I}_{t} \mbox{ and } \mathcal{I}_{t+1}$]{ \includegraphics[width=7.2cm]{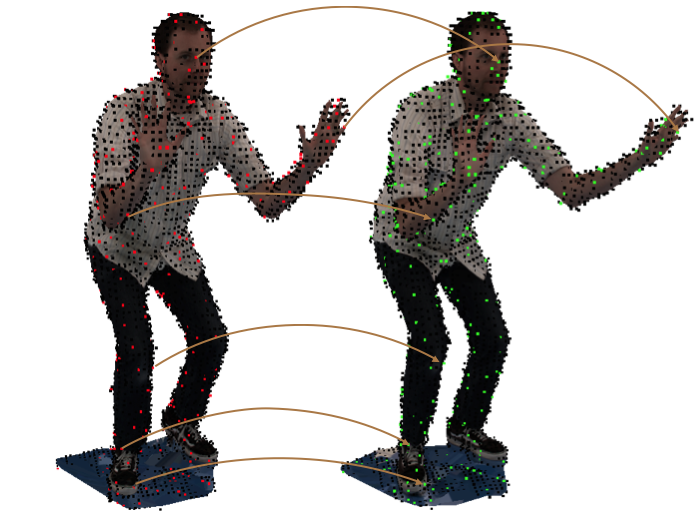} \label{man_sequence_matching}} 
            \subfigure[$ \mathcal{I}_{t,mc} +  \mathcal{I}_{t+1}$]{ \includegraphics[width=3.3cm]{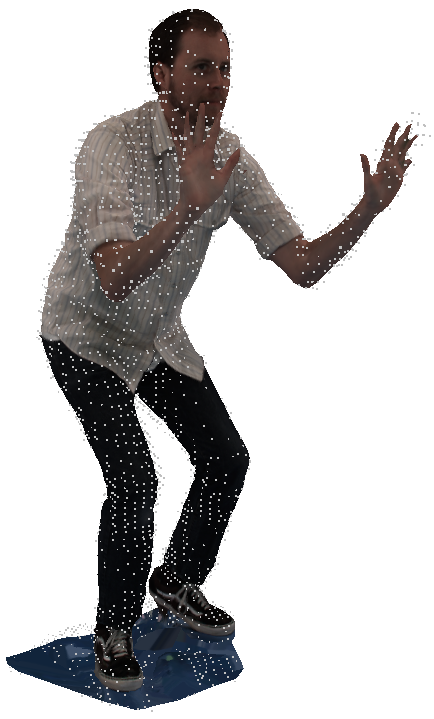}   \label{man_sequence_motion_compensated}}
           \caption{ Example of motion estimation and compensation in the yellow dress and the man sequence. The superimposition of the  reference ($\mathcal{I}_{t}$) and target frame ($\mathcal{I}_{t+1}$) is shown in  (a) and (d)  while in (b) and (e) we show the correspondences between the target (red) and the reference frame (green).  The superposition of the motion compensated reference frame ($\mathcal{I}_{t,mc}$) and the target frame ($\mathcal{I}_t$) is shown in (c), (f).  Each small cube corresponds to a voxel in the motion compensated frame. } 
        \label{fig:matching_examples_man}
\vspace{-0.1cm}
\end{figure*}

We illustrate in this section the matching performance obtained with our motion estimation scheme and the performance of the proposed compression scheme. We use  two different sequences that capture human bodies in motion, i.e., the yellow dress  and the man sequences, which have been voxalized to resemble data collected by  the real-time high resolution sparse voxalization algorithm \cite{Loop_2013}. The first sequence consists of 64 frames, and the second one of 30 frames.  For each frame, we voxelize the point cloud  to a voxel stepsize of 20, which generates a set of approximately 8500 occupied voxels out of a total of 75000 initial 3D points with color attributes. The 
 number of voxels  depends on the size of the actual frames.  As an illustrative example, in this paper the chosen voxel stepsize corresponds to an octree depth of seven. However, our motion estimation and compression scheme can be applied to any other octree level, with similar performance. The graph based interpolation problem of  (\ref{eq:closed_form}) is solved using MINRES-QLP \cite{Choi2014}. 

\subsection{Motion estimation}

We first illustrate the performance of our motion estimation algorithm by studying its effect in motion compensation experiments.  We select two consecutive frames for each sequence, namely the reference ($\mathcal{I}_t$) and the target frame ($\mathcal{I}_{t+1}$).  The graph for each frame is constructed as described in Section \ref{sec:representation_with_graphs}.  
We define spectral graph wavelets of  4 scales on these graphs, and for computational efficiency, we approximate them with Chebyshev polynomials of degree 30 \cite{Hammond2010}.   We select the number of representative feature points to be around 500,  which corresponds  to fewer than $10\%$ of the total occupied voxels,  and we compute the sparse motion vectors on the corresponding nodes by spectral matching.  
 We estimate the motion on the rest of the nodes by smoothing the motion vectors on the graph according to (\ref{eq:smoothing_eq}). 
 
In Figs. \ref{yellow_dress_example}, \ref{man_sequence_example}, we superimpose the reference and the target frame for the yellow dress and the man sequence accordingly  in order to illustrate the motion involved between two consecutive frames. The key points used for spectral matching in each of the two frames are shown in Figs. \ref{yellow_dress_matching}, \ref{man_sequence_matching} and they are represented in red for the target and in green for the reference frame. For the sake of clarity, we highlight only some of the correspondences used for computing motion vectors. We observe that the sparse set of matching vertices are accurate and well-distributed in space for both sequences.  Finally, in Figs. \ref{yellow_dress_motion_compensated},  \ref{man_sequence_motion_compensated}, we superimpose the target frame and the voxel representation of the motion compensated reference frame.  
By comparing visually these two figures to  \ref{yellow_dress_example}, \ref{man_sequence_example} respectively, we observe that in both cases the motion compensated reference frame is much closer to the target frame than the simple reference frame.  The obtained results confirm that our algorithm is able to estimate accurately the motion.

\subsection{3D geometry compression}
\begin{figure}[]
      \centering
          { \includegraphics[width=7cm]{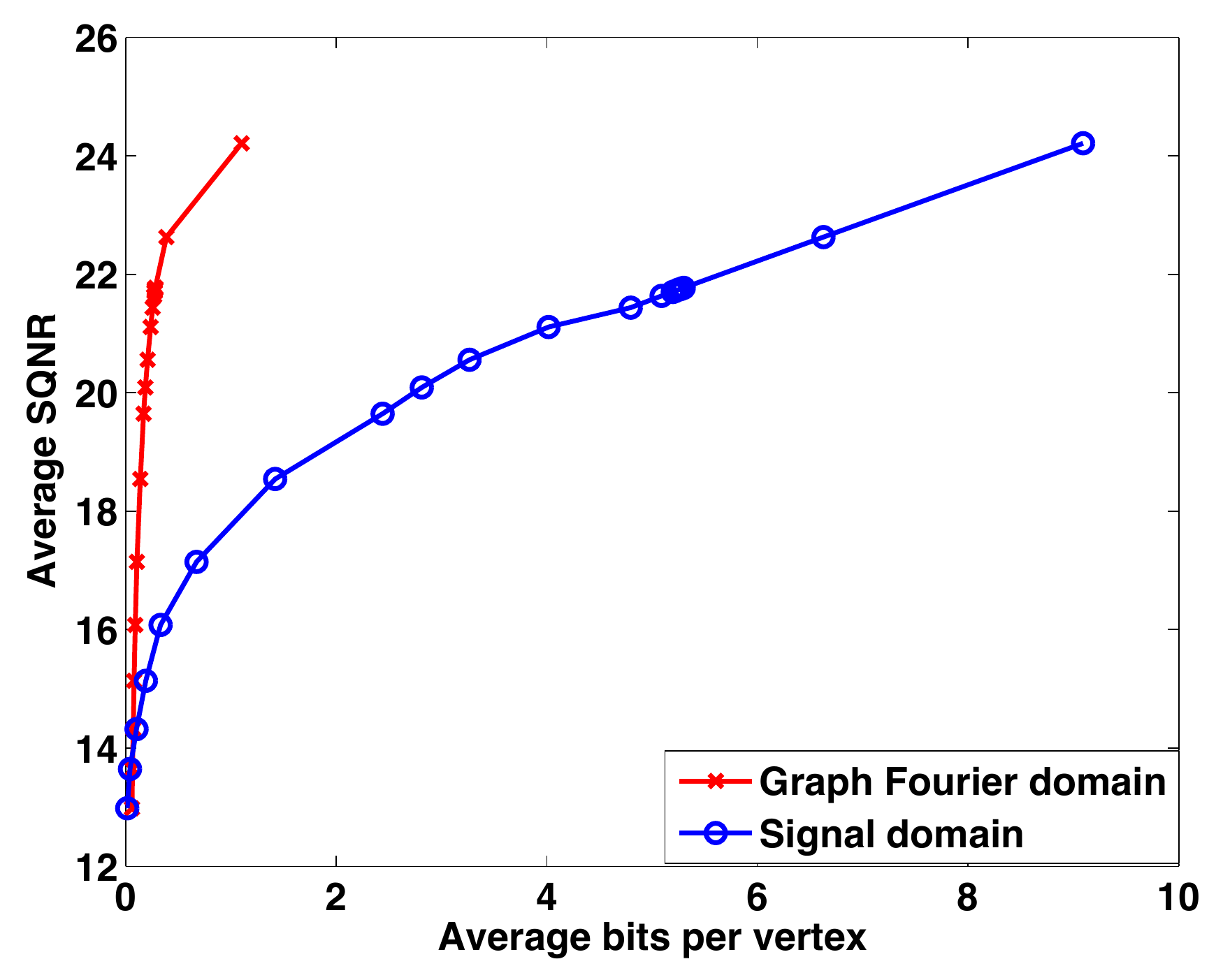} \label{motion_vectors_man}} 
           \caption{Performance comparison of the average signal-to-quantization noise ratio (SQNR) versus bits per vertex (bpv) for coding the motion vectors in the graph Fourier domain and in the signal domain. } 
        \label{fig:motion_vectors_man}
\vspace{-0.1cm}
\end{figure}

We now study the benefits of motion estimation in the compression of geometry in 3D point cloud sequences. The compressed geometry information includes motion vectors, the 3D positions of the reference frame, and the structural difference of the target frame and the motion compensated reference frame captured by the  XOR encoded information.     We note that the compression is performed on the whole sequence. The frames of the sequences are coded sequentially in the following way. Only the first frame is coded independently, while all the other frames are coded by using as a reference frame the previously coded frame.   We first code the motion vectors with the proposed coding scheme of Sec. \ref{subsec:motion_vector_coding}. The motion signal in each of the three directions is coded separately.   In Fig.~\ref{fig:motion_vectors_man} we show the advantage of transforming the motion vectors in the graph Fourier domain, in comparison to coding directly in the signal domain, for the man sequence.  Different stepsizes for uniform quantization are used to obtain different coding rates, hence different accuracies of the motion vectors.   The performance is measured in terms of the signal-to-quantization noise ratio (SQNR) for a fixed number of bits per vertex.  The SQNR is computed on pairs of frames. Each point in the rate distortion curve corresponds to the average over 64  frames. The results confirm that coding the motion vectors in the graph Fourier domain results in an efficient spatial decorrelation of the motion signals,  which brings significant gain in terms of coding rate.  Similar results hold for the yellow dress sequence, but we omit them due to the lack of space.

\begin{figure*}[]
      \centering
              \subfigure[Man sequence] { \includegraphics[width=7cm]{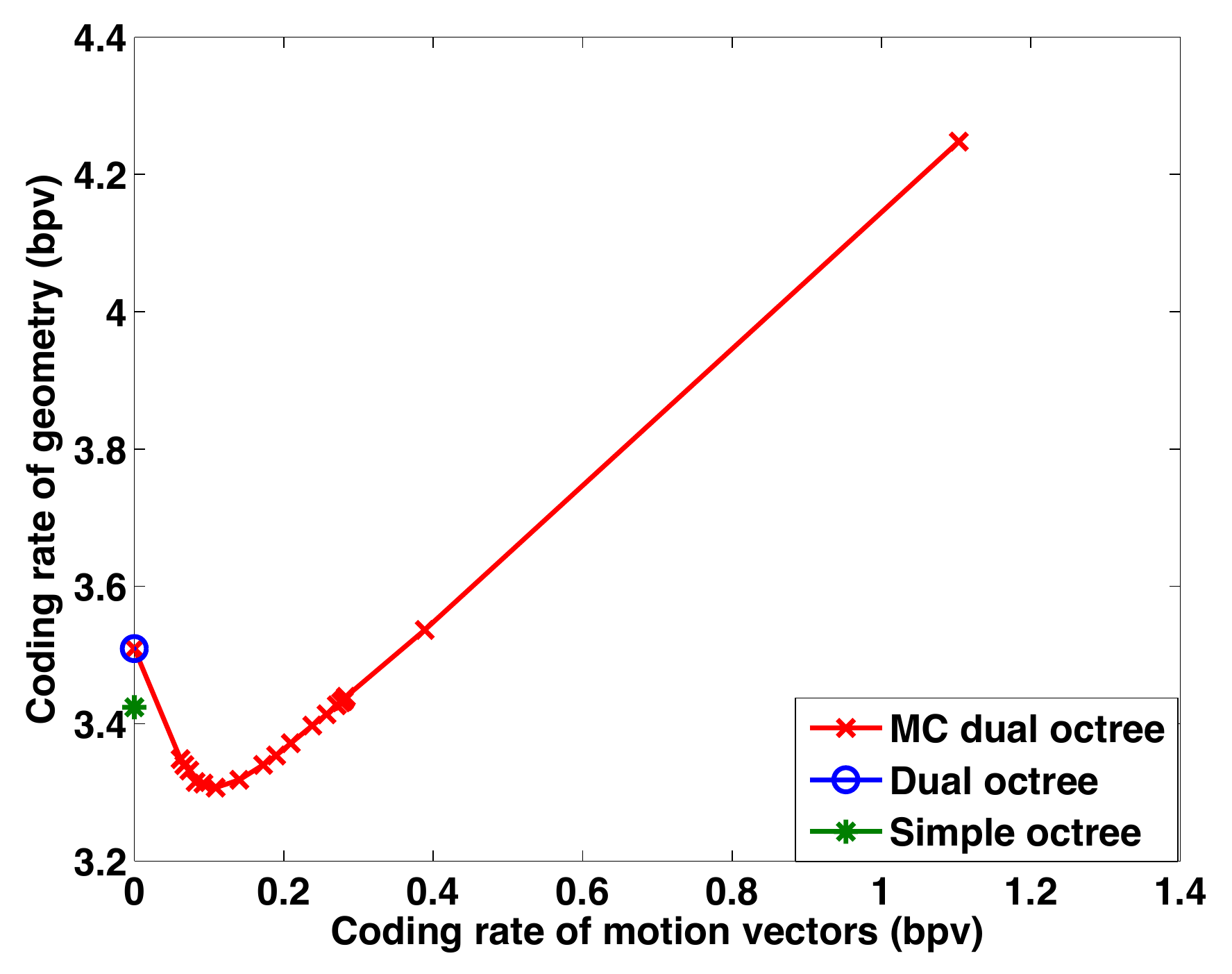} \label{geometry_compression_man}} 
            \subfigure[Yellow dress sequence]{ \includegraphics[width=7cm]{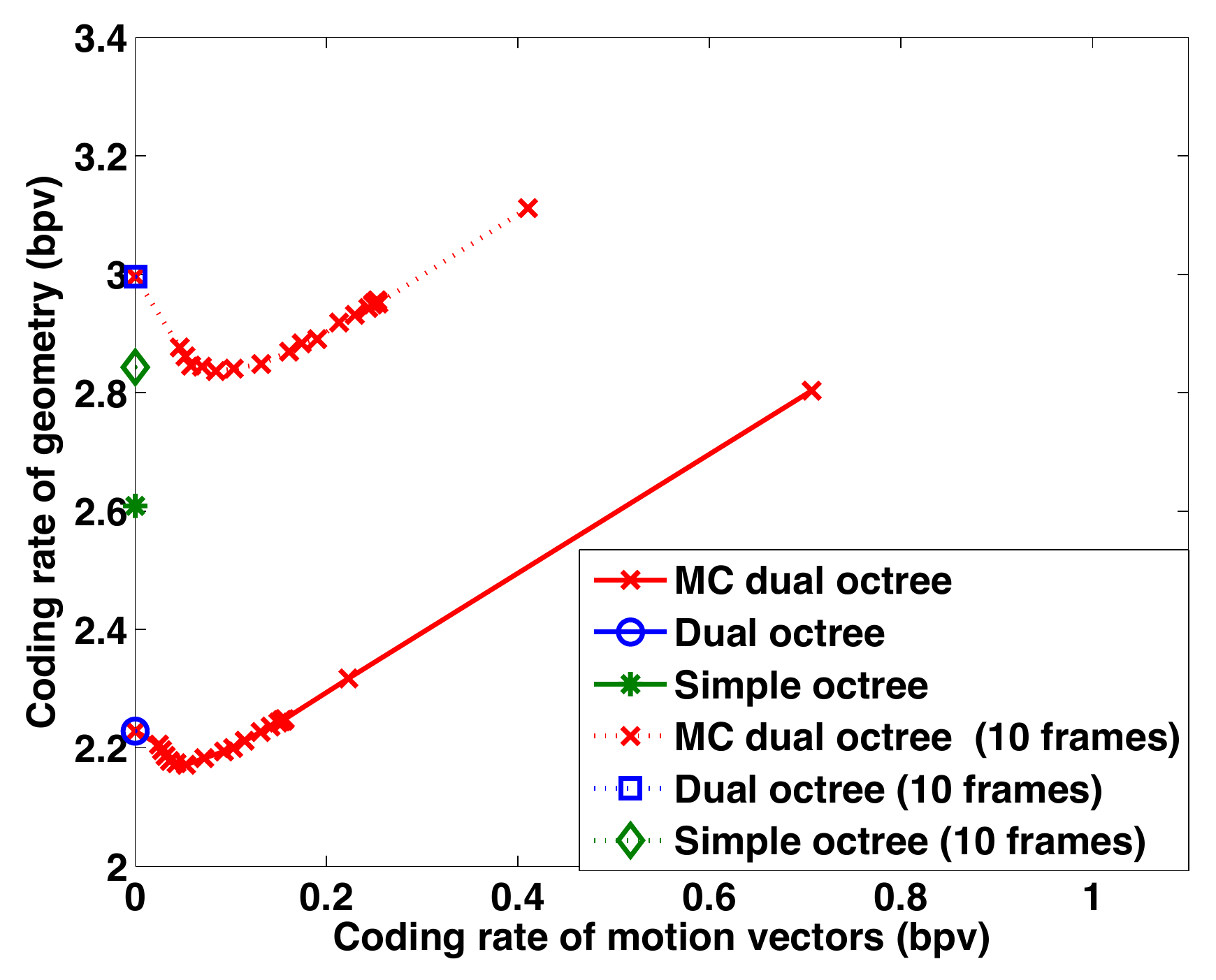} \label{geometry_yellow_dress}} 
           \caption{Effect of the coding rate of the motion vectors on the  overall coding rate of the geometry for the motion compensated dual octree algorithm. By sending the motion vectors at low bit rate ($\approx 0.1$ bpv), the motion compensated dual octree scheme outperforms slightly the simple octree and the dual octree compression algorithm. } 
        \label{fig:geometry_compression_man}
\vspace{-0.1cm}
\end{figure*}

We study next the effect of motion compensation in the coding rate of the 3D positions.   We recall that the coding of the geometry is lossless. There exists however a tradeoff between the overall coding rate of the geometry and the coding rate of the motion vectors as we illustrate next.   In particular, we compare the motion compensated dual octree scheme as described in Sec. \ref{subsec:geometry_coding}, to the dual octree scheme of  \cite{Kammerl2012}, and the simple octree compression algorithm.   In Fig.~\ref{fig:geometry_compression_man}, we illustrate the coding rate of the geometry with respect to  the coding rate of the motion vectors, measured in terms of the average number of bits per vertex (bpv) over all the frames, for each of the three competitive schemes.  The coding rate of the geometry includes the coding rate of the motion vectors.  In Fig.~\ref{geometry_compression_man},  the smallest coding rate of the geometry (3.3 bpv) for the man sequence is achieved for a coding rate of the motion vectors of only 0.1 bpv. The latter indicates that coarse quantization of the motion vectors is enough for an efficient geometry compression.  A smaller number of bits per vertex  however tends to penalize the effect of motion compensation,  giving an overall coding rate that  approaches the one of  the dual octree  compression scheme. Of course, a finer coding of the motion vectors increases the overhead in the total  coding rate of the geometry.  The corresponding numbers for the simple octree and the dual octree compression scheme are approximately 3.42 and 3.5 respectively. These results indicate that  the temporal structure captured by the dual octree compression scheme is not sufficient to improve the coding rate with respect to the simple octree compression algorithm. Motion compensation is thus needed to remove the temporal correlation.  However, the overall gain that we obtain is small and corresponds to $3.5\%$ and $5.7\%$  with respect to the simple octree and the dual octree compression algorithm respectively.   
  Similar results are observed in Fig.~\ref{geometry_yellow_dress} for the yellow dress sequence. In order to study the effect of the motion in the compression performance, we perform two different tests. In the first test, we compress the entire yellow dress sequence, which is a low motion sequence. In the second test, we sample the sequence by keeping only 10 frames that are characterized by  higher motion between consecutive frames. We then compress the geometry for this new smaller sequence.  In Fig.~\ref{geometry_yellow_dress}, we observe that when the motion is low, the motion compensated dual octree and the dual octree compression algorithms are much more efficient in coding the geometry in comparison to the simple octree compression algorithm. Moreover, the motion compensated dual octree scheme requires  a slightly smaller number of bits per vertex (2.2 bpv), for a coding rate of  the motion vectors of 0.1 bpv.     The coding rate for the dual octree  and the simple octree compression algorithm  are respectively 2.24 and 2.6 bpv. On the other hand, the simple octree compression scheme outperforms the dual octree compression algorithm, in the higher motion sequence of  10 frames,  with coding rates of 3 and 2.8 bpv respectively. The motion compensated dual octree compression algorithm can close the gap between these two methods by achieving a coding rate of 2.8 bpv. We note that this performance is achieved for an  overhead of 0.15 bpv for coding the motion vectors.   Due to this overhead, the performance of the simple octree and the motion compensated dual octree compression algorithm are relatively close. To summarize, the above results indicate that, although motion compensation can improve the overall geometry compression performance, this improvement is marginal.

\subsection{Color compression}

In the next set of experiments, we use  motion compensation  for color prediction,  as described in Section \ref{subsec:color_coding}. That is, using the smoothed motion field, we warp the reference frame $\mathcal{I}_{t}$ to the target frame $\mathcal{I}_{t+1}$, and predict the color of each point in $\mathcal{I}_{t+1}$ as the average of the three nearest points in the warped frame $\mathcal{I}_{t,mc}$.   We fix the coding rate of the motion vectors to 0.1 bpv and, for the sake of comparison, we compute the signal-to-noise ratio (SNR) after predicting the color in the following  three different ways:   (i)   the colors of points in the target frame are predicted from their nearest neighbors in the warped frame $\mathcal{I}_{t,mc}$,  (ii)  the colors of points in the target frame are predicted from their nearest neighbors in $\mathcal{I}_{t}$, and (iii) the colors of points in the target frame are predicted as the average color of all the points in $\mathcal{I}_{t}$. The SNR for frame $\mathcal{I}_{t+1}$ is defined  as $\mbox{SNR}_{t+1} = 20 \log_{10}\frac{\|c_{t+1}\|}{\|c_{t+1} - \widetilde{c_{t+1}}\|}$, where we recall that $c_{t+1}$ and $\widetilde{c_{t+1}}$ are the actual color and the color prediction respectively. The prediction error is measured by taking pairs of frames in the sequence and computing the average over all the pairs.  The obtained values for the man sequence are (i) 13 dB, (ii) 10.5 dB, and (iii) 4 dB, while for the yellow dress sequence the corresponding values are (i) 17 dB, (ii) 15 dB, and (iii) 6.5 dB. We notice that for both sequences motion compensation can significantly reduce the prediction error, by obtaining an  average gain  in the color prediction of 2.5 dB and 8-10 dB   with respect to predicting simply based on the color of the nearest neighbors in the reference frame, and the average color of the reference frame respectively.

\begin{figure}[]
      \centering
         
            { \includegraphics[width=7cm]{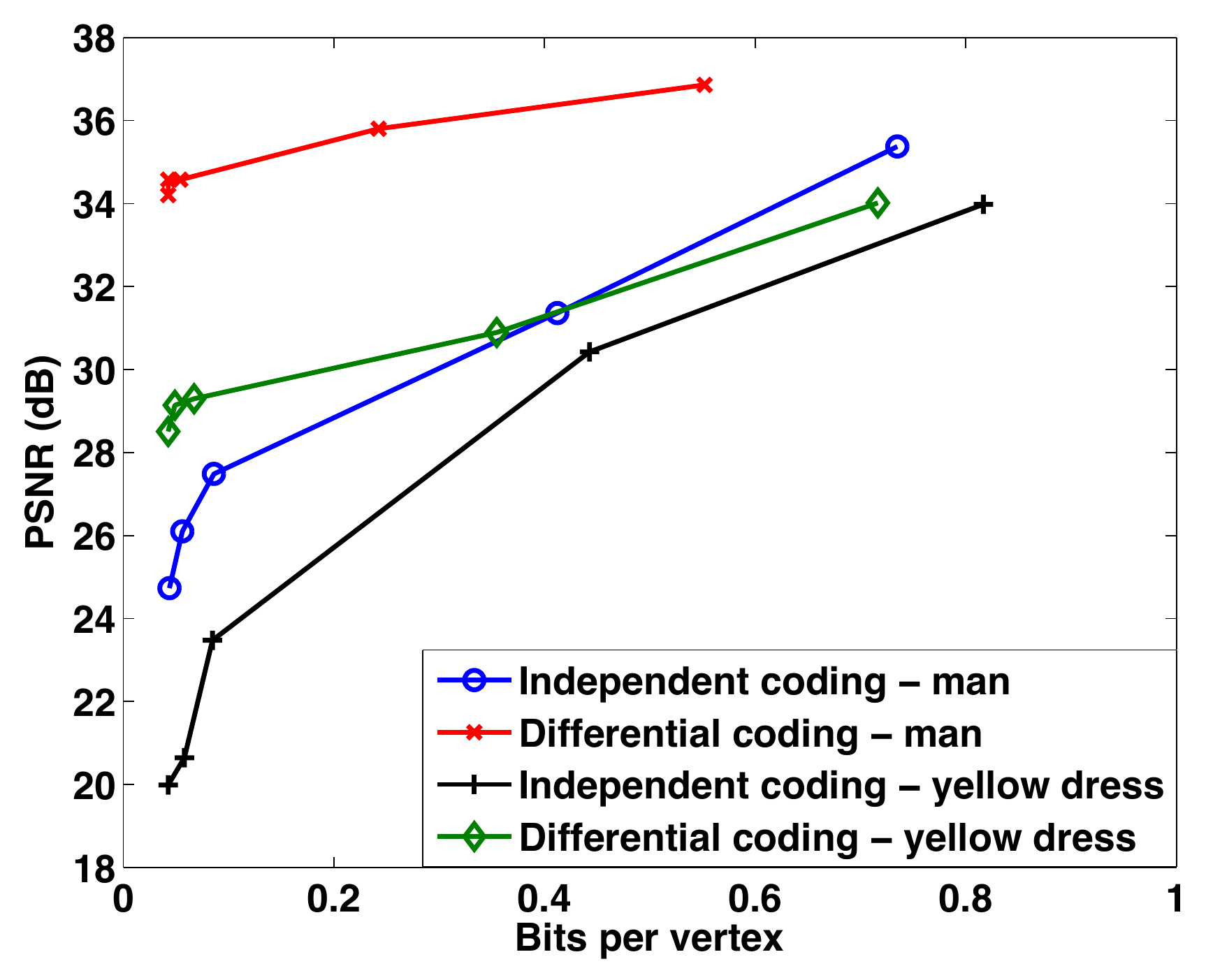} \label{color_compression_man}} 
           \caption{Compression performance (dB) vr. bits per vertex for independent  and differential coding  on both datasets for a quantization stepsize of $\Delta = [32,~ 64,~ 256,~ 512,~ 1024]$.  } 
        \label{fig:compression_comparison}
\vspace{-0.1cm}
\end{figure}

We finally use the prediction obtained from our motion estimation and compensation scheme to build a full scheme  for color compression, that is based on a prediction path of a series of frames.   Compression of color attributes is obtained by coding the residual of the target frame with respect to the color prediction obtained with the  scheme described in Section \ref{subsec:color_coding}.  
In our experiments, we code the color in small  blocks  of $16\times16\times16$ voxels. We measure the PSNR obtained for different levels of the quantization stepsize in the coding of the color information, hence different coding rates,  for both independent \cite{Zhang2014} and differential coding. The results  for both datasets are shown in Fig.~\ref{fig:compression_comparison}. Each point on the curve corresponds to the average PSNR of  the R,G,B components across the first ten frames of  each sequence, obtained for a quantization stepsize of $\Delta= [32, 64, 256, 512, 1024]$ respectively.  We observe that at low bit rate ($\Delta = 1024$), differential coding provides a  gain with respect to independent coding of  approximately 10 dB for both sequences. On the other hand, at high bit rate, the difference between independent and differential coding tends to become smaller, as both methods can code the color quite accurately.  We note that the gain in the coding performance  is highly dependent on the length of the prediction path.  As the number of predicted frames increases, the accumulated quantization error from the previously coded frames is expected to lead to a gradual PSNR degradation that is more significant at low bit rate.  This can be mitigated by periodic insertion of reference frames, and by optimizing the number of predicted frames between consecutive reference frames.

\subsection{Discussion}
 Motion compensation is beneficial overall in the compression of 3D point cloud sequences. The main benefit though is observed in the coding of the color attributes, providing a gain of up to 10 dB with respect to coding each frame independently. The gain in the compression of the 3D geometry is only marginal due to the overhead for coding the motion vectors. Finally, from the experimental validation in both datasets, we observe that from the overall bit budget the most significant part is used for the compression of the geometry. On the other hand, a very coarse quantization of the motion vectors is sufficient to achieve an overall good compression rate.  For example, for each vertex in the man sequence, we need 0.1-0.2 bits to code the motion vectors, 0.1-0.3 bits for the color residual, and 3.3 bits for the geometry compression. Similar observations hold for the second dataset.  The proposed motion compensated geometry compression framework that is based on the differential coding of consecutive octree graph structures is the most expensive part of the overall compression system. For a particular depth of the tree, the compression is lossless.  A lossy geometry compression scheme that reduces the coding rate  seems to be an interesting future direction.

\section{Conclusions}
\label{sec:conclusions}
In this paper, we have proposed a novel compression framework for 3D point cloud sequences that is based on exploiting temporal correlation between consecutive point clouds. We have first proposed an algorithm for motion estimation and compensation.   The algorithm  is based on the assumption that  3D models are representable by a sequence of weighted and undirected graphs and the geometry and the color of each model can be considered as graph signals residing on the vertices of the corresponding graphs. Correspondence between a sparse set of nodes in each graph is first determined  by matching descriptors based on spectral features that are localized on the graph. The motion on the rest of the nodes is interpolated by exploiting the smoothness of the motion vectors on the graph. Motion compensation is then  used to perform geometry and color prediction. Finally, these predictions are used to differentially encode both the geometry and the color attributes. Experimental results have shown that the proposed method is efficient in estimating the motion and it eventually provides significant gain in the overall compression performance of the system.

\section*{\textsc{Acknowledgements}}
The authors would like to thank Cha Zhang and Dinei  Flor\^{e}ncio for providing the code  for  the color encoder and  help in  the color compression experiments.

\bibliographystyle{IEEEbib}

\bibliography{mybibfile}

\end{document}